\documentclass[10pt, a4paper]{article}
\usepackage{lrec2022} % this is the new LREC2022 Style
\usepackage{multibib}
\newcites{languageresource}{Language Resources}
\usepackage{graphicx}
\usepackage{tabularx}
\usepackage{soul}
\usepackage{titlesec}
\titleformat{\section}{\normalfont\large\bfseries\center}{\thesection.}{1em}{}
\titleformat{\subsection}{\normalfont\SmallTitleFont\bfseries\raggedright}{\thesubsection.}{1em}{}
\titleformat{\subsubsection}{\normalfont\normalsize\bfseries\raggedright}{\thesubsubsection.}{1em}{}
\renewcommand\thesection{\arabic{section}}
\renewcommand\thesubsection{\thesection.\arabic{subsection}}
\renewcommand\thesubsubsection{\thesubsection.\arabic{subsubsection}}
\usepackage{epstopdf}
\usepackage[utf8]{inputenc}

\usepackage{hyperref}
\usepackage{xstring}

\usepackage{color}

\usepackage{paralist}
\usepackage{cleveref}
\usepackage{enumitem}

\setlist[itemize]{noitemsep}
\setlist[enumerate]{noitemsep}
\title{Klexikon: A German Dataset for Joint Summarization and Simplification}

\name{Dennis Aumiller and Michael Gertz} 

\address{Institute of Computer Science, Heidelberg University \\
         Im Neuenheimer Feld 205, Heidelberg, Germany \\
         \{aumiller, gertz\}@informatik.uni-heidelberg.de\\}

\abstract{
Traditionally, Text Simplification is treated as a monolingual translation task where sentences between source texts and their simplified counterparts are aligned for training.
However, especially for longer input documents, summarizing the text (or dropping less relevant content altogether) plays an important role in the simplification process, which is currently not reflected in existing datasets.
Simultaneously, resources for non-English languages are scarce in general and prohibitive for training new solutions.
To tackle this problem, we pose core requirements for a system that can jointly summarize and simplify long source documents.
We further describe the creation of a new dataset for joint Text Simplification and Summarization based on German Wikipedia and the German children's encyclopedia "Klexikon", consisting of almost $2{,}900$ documents.
We release a document-aligned version that particularly highlights the summarization aspect, and provide statistical evidence that this resource is well suited to simplification as well. Code and data are available on Github: \url{https://github.com/dennlinger/klexikon}
 \\ \newline \Keywords{Summarization, Text Simplification, German} }
\begin{document}

\maketitleabstract

%%%%%%%%%%%%%%%%%%%%%%%%%%%
\section{Introduction}

The goal of Text Simplification (TS) is to produce easily understandable texts that benefit disadvantaged readers such as children, dyslexic, or language learners. Simplifications are often generated by adapting a source text written for adult/native readers. 
However, recent work in simplification has mostly addressed TS as a monolingual translation task, where individual sentences are "translated" into a simplified version \cite{zhu-etal-2010-monolingual,coster-kauchak-2011-simple,hwang-etal-2015-aligning}.
The main focus is put on either lexicographic replacements, paraphrasing, sentence splitting, or the dropping of words within a single sentence \cite{amancio-specia-2014-analysis}, which implies that the simplification of any input document will consist of roughly the same number of sentences.
While this approach is appropriate for sufficiently short source documents, longer articles become strenuous for disadvantaged readers. As can be seen in \Cref{tab:corpusstats}, articles in different corpora come with varying lengths of their respective source texts. When simplifications are generated via manual sentence-by-sentence translations, the simplified texts tend to have more sentences than the source documents. When alignments are constructed from a source and simplification text on the same topic instead, they exhibit a drastic length disparity.

Current simplification systems are, however, inherently limited in their ability to address the problem of joint simplification and summarization from much longer input documents.
Sentence-level alignments were traditionally seen as one way to circumvent certain problems in TS, namely:
\begin{enumerate}
	\item Human feedback for judging simplification quality is more consistent for sentences, compared to longer samples, such as entire documents.
	\item Metrics such as BLEU \cite{papineni-etal-2002-bleu} or SARI \cite{xu-etal-2016-optimizing} rely on (aligned) reference texts for automated evaluation.
	\item Prior alignment of sentences limits the length of input samples, which is essential for algorithms with non-linear runtime, or length constraints.
\end{enumerate}

\begin{table}[t!]
	\centering
	\begin{tabular}{l@{\hspace{-0.8em}}r|r|r}
		\hline
		& \textbf{Aligned} & \multicolumn{2}{c}{\textbf{Avg.~\#Sentences}} \\
		\textbf{Resource} & \textbf{Articles} & \textbf{Source} & \textbf{Simple}\\
		\hline
		Klexikon (Ours) & $2{,}898$ & $242.09$ & $32.51$ \\
		\cite{hewett-stede-2021-automatically} & $978$ & $10.12$ & $43.54$ \\
		\cite{battisti-etal-2020-corpus}$^*$ & $378$ & $45.29$ & $55.75$ \\
		\hline
		\cite{kauchak-2013-improving} & $59{,}775$ & $64.52$ & $8.46$ \\
		\cite{xu-etal-2015-problems}$^*$ & $1{,}130$ & $49.59$ & $51.27$ \\
		
	\end{tabular}
	\caption{%Number of documents and average number of sentences from document-aligned resources with normal and simplified texts. 
		Corpus statistics for datasets with document alignments in German (top) and English (bottom). $^*$ indicates resources created by simplifying articles sentence-by-sentence. For \protect\cite{xu-etal-2015-problems,hewett-stede-2021-automatically}, we refer to the respective simplified corpora with simplification level 1.}
	\label{tab:corpusstats}
\end{table}

In this work, we present remedies to the problem of missing document alignments, and argue that the inclusion of summarization into the broader context of Text Simplification is a necessary step towards end-to-end solutions for longer input texts. Specifically, it addresses the following problems:
\begin{enumerate}
	\item Long-form documents can be compressed into significantly shorter summarized simplifications.
	\item Document alignments provide context for models that are otherwise based on single sentence pairs.
	\item The amount of accessible training data increases, which is especially important for languages other than English, where data is generally scarce.
\end{enumerate}

Simultaneously, TS offers interesting challenges to the summarization community, which hopefully facilitates exchange between the two fields: 
On existing summarization datasets, simply taking the leading three sentences offers strikingly good results \cite{nallapati2017summarunner}, which may lead to systems learning specific extractive strategies instead of generalizing to broader textual relevance. Preliminary experiments show that our dataset poses a harder challenge for summarization systems, due to the additional simplification aspect.

Our proposed resource was obtained from semi-automated alignments between the German Wikipedia and the children's encyclopedia "Klexikon" \cite{schulte-dijk-2015-free}, written for children aged 8 to 13 years.\footnote{\url{https://klexikon.zum.de/wiki/Hilfe:Grunds\%C3\%A4tze}, accessed: 15.01.2022}
With almost $2{,}900$ articles, it is the largest non-English resource with document alignments.

%%%%%%%%%%%%%%%%%%%%%%%%%%%
\section{Related Work}
Related work can broadly be categorized into relevant simplification work, and associated works on resources for (German) summarization datasets.

\subsection{Text Simplification}
Previously mentioned work frequently deals with data aligned based on Simple Wikipedia \cite{zhu-etal-2010-monolingual,coster-kauchak-2011-simple,hwang-etal-2015-aligning}. The main differences between these approaches lie in their alignment strategies and underlying simplification model.
The only work on Simple Wikipedia that specifically introduces a document-aligned version is \cite{kauchak-2013-improving}, who investigates performance gains from supplementing language models with additional (non-simplified) texts. Importantly, it is not explicitly used for learning simplification.
%\cite{glavas-stajner-2015-simplifying} go as far as suggesting an unsupervised system without the necessity of aligned corpora at all.

\cite{hancke-etal-2012-readability} introduced a first German resource containing simplified texts based on unaligned articles from GEO and GEOlino, a German magazine similar to National Geographic, and its edition specifically for children.
They build a classification system that is able to classify between normal and simplified texts for several article categories.
A larger and improved version from the same source was collected by \cite{weiss-meurers-2018-modeling}, who also introduce a resource based on transcripts from German TV broadcasts (Tagesschau/Logo!), again without any alignment.
%However, neither resource is document-aligned, and therefore not suitable for generating document-level simplifications.
The first mention of an aligned corpus for German can be found in \cite{klaper-etal-2013-building}, who automatically align websites with their corresponding versions in accessible language. Their corpus contains a total of about 270 articles.

Most recently, \cite{battisti-etal-2020-corpus} collected a larger corpus, where 378 texts contain document alignments. Arguably, unaligned resources might still be helpful to facilitate pre-training of models.
In an attempt to circumvent data scarcity, \cite{mallinson-etal-2020-zero} employ multi-lingual pre-training, which they tested with a small, manually labeled German evaluation set.

To our knowledge, \cite{hewett-stede-2021-automatically} were the first to utilize alignments between Wikipedia and Klexikon, with an additional extension to MiniKlexikon, a secondary simplification level. Due to the further required alignments, the overall size of their data is about 10\% of our presented corpus. To avoid problems stemming from extreme length discrepancies, they also only extract introduction and abstracts for Wikipedia articles, which is something we explicitly encourage in our version. This also explains the different lengths while using the same document sources, as reported in \Cref{tab:corpusstats}~.

\subsection{Summarization}
\cite{parmanto2005access} are the first to explicitly explore summarization and simplification in a common context, albeit for the task of website accessibility.
Further work models summarization itself as a simplification technique, e.g., 
\cite{margarido2008automatic} investigated extractive summarization approaches and how they help disadvantaged readers. A similar experiment was conducted by \cite{smith-jonsson-2011-automatic} for Swedish texts, who find summarized texts to be more readable as well.
Also dealing with extractive summarizers, \cite{finegan2016sentence} look at simplifications in the biomedical and legal domain, but their findings indicate that altered sentences lead to fewer correctly answered questions by domain experts.
Simplification has also been suggested for multi-document summarization:
\cite{siddharthan-etal-2004-syntactic} select relevance exclusively over syntactically simplified sentences, whereas other works use simplification as an alternative to regular sentence selection \cite{vanderwende2006beyond,yih2007multi}.

Closest to a unified framework is the work by \cite{ma2017semantic}, who use the same neural encoder-decoder architecture for separate simplification and summarization tasks, which highlights the shared similarities in terms of shared model architectures and training.

To our knowledge, there exist few resources for German single-document summarization. \cite{nitsche2019abstractive} mention a (private) resource, provided by the German Press Agency (dpa), which uses headlines as target summaries. \cite{frefel-2020-summarization} generate a corpus based on German Wikipedia articles, and treat the overview paragraph at the beginning as the summary of the article. A similar approach including cross-lingual alignments between English and German has also been recently published \cite{fatima-strube-2021-novel}~.

%%%%%%%%%%%%%%%%%%%%%%%%%%%
\section{Text Simplification with Joint Summarization}

\begin{figure*}[ht]
	\centering
	\includegraphics[width=0.32\textwidth]{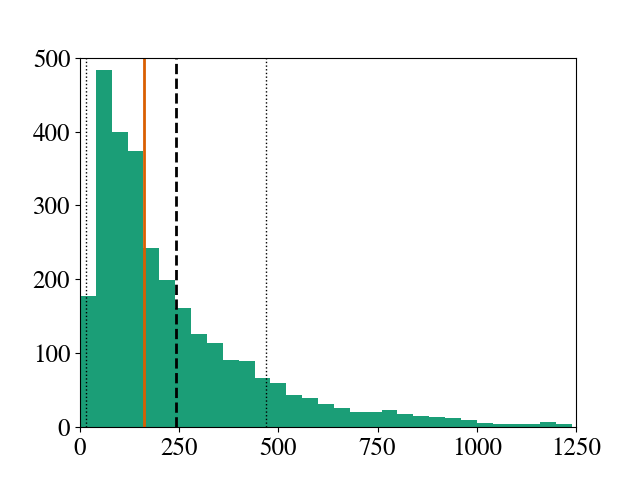}
	\includegraphics[width=0.32\textwidth]{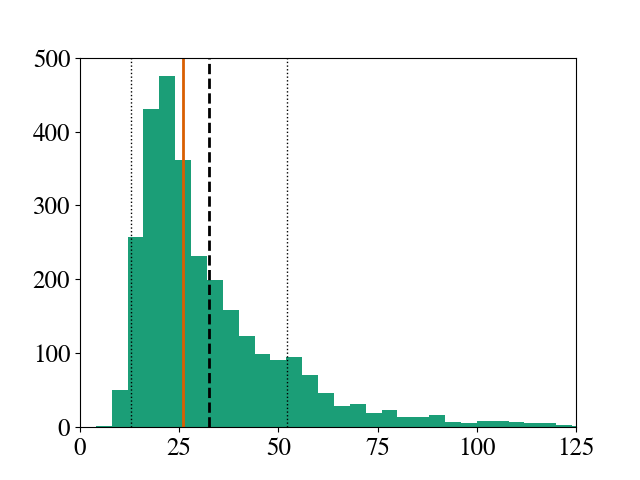}
	\includegraphics[width=0.32\textwidth]{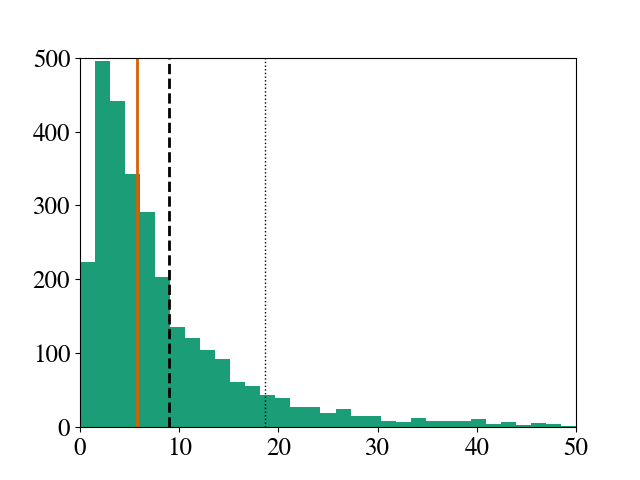}
	\caption{Histogram of our Klexikon dataset by number of sentences. Displayed are the distribution for source texts (left; bin width 50), simplified articles (center; bin width 5), and compression ratio of source over simplified lengths (right; bin width 2). Vertical lines represent median length (continuous orange), mean length (dashed black) and one standard deviation (dotted black).}
	\label{fig:stats}
\end{figure*}

As previous work has shown, summarization in itself can already be considered a weaker form of simplification \cite{margarido2008automatic,smith-jonsson-2011-automatic}, although existing work never formalizes TS as a summarization problem.
Several points have to be addressed by both simplification and summarization components for a full end-to-end solution. In this section, we outline suggestions for a unified system design.

\subsection{Considerations for Simplification}
As previously stated, current simplification systems cannot generate significantly shorter output texts when simplifying individual sentences. 
This is mainly due to the sentence-aligned training setup instead of training with the entire input document. 
Further, this drops a sizable portion of the source text from training, since sentences are only considered when they align directly with a simplified part.
Several resources also lack a document alignment altogether, which completely precludes them from being used as a training resource for end-to-end systems.

Importantly, relevance of individual segments (sentences or paragraphs) has to be computed \emph{without knowledge about the output corpus}.
This can, for example, be achieved by pre-training strategies on monolingual corpora \cite{mallinson-etal-2020-zero}, but could otherwise be learned as an intermediate step in neural architectures.
This has been previously shown to work well for multi-document summarization \cite{liu-lapata-2019-hierarchical}, where paragraph relevance was learned across several documents.

Further, existing manually annotated corpora are frequently generating simplifications of short texts by "translating" sentence-by-sentence. This reinforces the bias towards equally long documents, which cannot be observed in post-aligned resources (i.e., where existing simplified texts were written independently on the same topic, cf. \Cref{tab:corpusstats}).
An amended assumption is that simplifications may only be \emph{up to a certain length}, due to varying attention spans of the target groups. This then requires additional "simplification" based on the length of the source document.
This could also be used as a parameter to model levels of difficulty, which is available for some resources, see the Newsela corpus \cite{xu-etal-2015-problems}.

Lastly, existing evaluation metrics strictly focus on sentence-level references \cite{xu-etal-2016-optimizing}. Extending system evaluations to document-level simplifications poses challenges that need to be overcome in order to collect both manual and automated feedback on the simplification quality.

\subsection{Considerations for Summarization}
For summarization, TS offers additional challenges not considered by current works.
Given a high enough compression rate, simplification can be seen as a special case of summarization. However, existing metrics, such as  ROUGE \cite{lin-2004-rouge}, rely on the re-appearance of $n$-grams in the target summary (in our case, the simplification). This is not guaranteed, given that the simplification can appear in the form of lexicographic replacements. 
It is thus unclear whether simplification should be considered a separate criterion or jointly modeled for the evaluation of summaries, specifically when considering other input factors as well \cite{ter2020makes}.
Additionally, the varying vocabulary and sentence structure pose a challenge to summarization systems, especially extractive approaches. See \Cref{sec:baseline} for experiments on our Klexikon corpus.
Previous work in that direction has mostly dealt with sentence-level lexicographic simplifications \cite{siddharthan-etal-2004-syntactic}, yet there are several other simplification operations to be considered \cite{amancio-specia-2014-analysis} in a joint end-to-end system.

%\subsection{Why is this interesting for Simplification?}
%\begin{itemize}
%	\item Already mentioned some of the points 
%	\item It is the largest document-aligned resource for German (Geolino has some rivaling sources, but discards a lot when looking at sentence-level alignment. Also, I'm still not sure if they align by document, but we are quite sure of our alignment).
%	\item The Geolino corpus isn't publicly available (only upon request).
%	\item We still lack end-to-end solutions for simplification tasks, which can take arbitrary inputs and produce simplified outputs, that don't operate on the sentence-level (paragraph context, topical context, etc.)
%	\item We can still go ahead and compute alignments with something like CATS to have a ''classic'' resource available.
%	
%\end{itemize}

%\subsection{Why is this interesting for Summarization?}
%\begin{itemize}
%	\item Many of the existing corpora focus on fixed topics (e.g., CNN/Daily mail with news articles). This is not very useful for generalization.
%	\item German has very few datasets available for summarization, basically only the AIPHES stuff, and a Master Thesis mentions a DPA dataset, which I can't even find online...
%	\item Compared to other simplification datasets, we specifically select articles that have a minimum degree of difference in the length of respective source/target pairs. We average a 8x longer Wiki article, compared to the simplified ''summary''.
%\end{itemize}

%%%%%%%%%%%%%%%%%%%%%%%%%%%
\section{Klexikon Dataset}

We introduce a new dataset, loosely inspired in its construction by English Simple Wikipedia, to facilitate future research in joint simplification and summarization.
Specifically, we use the German children's encyclopedia "Klexikon" to obtain simplifications, and align them with reference articles from the German Wikipedia.
Compared to Simple Wikipedia, which can be freely edited, Klexikon specifically targets children between roughly the age of 8-13 as readers, and follows a strict reviewing procedure for individual articles, resulting in higher quality texts.
We only consider Wikipedia articles with a minimum length of 15 paragraphs, which helps to filter out disambiguation pages or stubs´. Additionally, this results in a clear contrast in overall article length between source and simplified texts (cf. \Cref{tab:corpusstats} and \Cref{fig:stats}).
The final dataset consists of $2{,}898$ article pairs, with Wikipedia documents having on average $8.94$ times more sentences compared to their Klexikon counterparts.
%for which we provide an overview of more statistics in \Cref{tab:detailed}.

%%%%%%%%%%%%%%%%%%%%%%%%%%%%%%%%%%%%%%
\subsection{Corpus Creation}
All manual steps during corpus creation were performed by the first author of this work.
We begin the extraction based on the list of all available articles from the Klexikon overview page in April 2021\footnote{\url{https://klexikon.zum.de/wiki/Kategorie:Klexikon-Artikel}, accessed 14.04.2021}~. At the time of experimentation, this returned 3{,}150 Klexikon articles, although more articles have been added since\footnote{In April 2022, the number of available articles has increased to 3{,}269.}.

\subsubsection{Document Alignment Strategy}
For the identification of matching articles between German Wikipedia and Klexikon, the following steps were performed:
\begin{enumerate}
	\item Querying the MediaWiki Search API\footnote{\url{https://www.mediawiki.org/wiki/API:Search}, accessed: 14.04.2021} with the title of the Klexikon article.
	2{,}861 articles, or around 90\%, have an entry with a directly matching heading on Wikipedia.
	However, this may include disambiguation pages or stubs.
	\item All remaining 289 unmatched articles are manually matched against the top five suggestions by the Wikimedia Search API.
	If no candidate article is appropriate, the entry is dropped from the corpus.
	\item Wikipedia articles with less than 15 paragraphs (108 articles) are again flagged and manually reviewed. Short Wikipedia entries may correspond to disambiguation pages (see next step), or are otherwise dropped because of their short length.
	\item Disambiguation pages are replaced with a specific Wikipedia page, if it topically matches at least 66\% of the Klexikon paragraphs.
	\footnote{For example, the Klexikon article for "Adler" (eagle) primarily talks about the animal, which is then chosen as the corresponding page in Wikipedia.}
\end{enumerate}

\subsubsection{Text Extraction}
The Klexikon website runs on the Wiki software, which makes text extraction across platforms very similar. For both websites, we extract all direct children elements of the main content block (div-class: \texttt{mw-parser-output}).
Of those, we only use text within \texttt{<p>} tags as the main paragraph content, and heading elements \texttt{<h1>}-\texttt{<h5>}. 
This simultaneously discards non-textual contents, e.g., images, as well as malformed text elements, such as image captions or lists.
We note that the removal of lists can also remove valid content, but frequently suffers from inconsistent grammatical correctness; while some bullet lists are equivalent to a self-contained paragraph, more often than not, it simply contains enumerations.

Further limitations for summarization include the potential content split on Wikipedia. For example, in the Klexikon article about the city of \emph{Aarhus}, there is explicit information about the ARoS (Aarhus art museum); however, on Wikipedia, this information would be found in the article about the \emph{museum itself}, and not in the page about the city. For now, we defer these edge cases to future extensions including multi-document summarization/simplification. 

To avoid encoding errors, we drop any character that appears less than 100 times in the corpus; more frequently appearing special characters are mapped to the closest latin character (e.g., \emph{\'{a}} to \emph{a}), with the exception of \emph{äöüß}, which are part of the standard German alphabet. In the absence of a close mapping (e.g., for Cyrillic letters), the character is dropped as well.
This assumes that foreign characters are irrelevant for simplified texts, which we can indeed observe from the utilized character set in Klexikon articles.
We process the raw text with spaCy's\footnote{\url{https://spacy.io}~, version 3.2} \texttt{de-core-news-md} model to separate sentences.
Our final data format maintains the following document representation:
\begin{enumerate}
	\item Line-by-line sentence representations based on spaCy boundary detection,
	\item Additional indication of separation of paragraphs (original \texttt{<p>} elements), and
	\item Highlighted headings according to the indicated level (heading, subheading, etc.), available primarily for the Wikipedia documents.
\end{enumerate}
A statistical view of the corpus can be found in \Cref{tab:detailed}~.

\begin{table}[t]
	\centering
	\begin{tabular}{lr|r}
		\hline
		\textbf{} & \textbf{Wikipedia} & \textbf{Klexikon} \\
		\hline
		Documents & $2{,}898$ & $2{,}898$ \\
		\hline
		Average sentences & $242.09$ & $32.51$ \\
		SD sentences & $227.39$ & $19.73$ \\
		Median sentences & $162$ & $26$ \\
		\hline
		Average tokens & $5{,}442.83$ & $436.87$ \\
		SD tokens & $5{,}093.82$ & $270.00$ \\
		Median tokens & $3{,}705$ & $347$ \\
		
	\end{tabular}
	\caption{Corpus statistics of the Klexikon dataset. SD refers to one standard deviation.}
	\label{tab:detailed}
\end{table}

\subsubsection{Sentence Alignments}
We also experimented with the creation of an automatically sentence-aligned variant of our data set.
Unfortunately, existing alignment algorithms from the TS community are not applicable here. CATS~\cite{stajner-etal-2018-cats} is one representative from the class of greedy alignment algorithms; these base their alignments on the assumption that a similar order of the content exists for both the source and simplification texts. This does not apply to our dataset, since texts have been written independently.
Algorithms with non-greedy alignment strategies exist~\cite{paetzold-etal-2017-massalign,jiang-etal-2020-neural}, but lack compatibility with German texts.

We instead experimented with alignments based on sentence embeddings from sentence-transformers~\cite{reimers-gurevych-2019-sentence}\footnote{\texttt{paraphrase-multilingual-mpnet-base-v2}, a multilingual variant also suitable for German texts.}, and selecting the most similar source sentence (or pair of sentences) for each Klexikon sentence.

However, sentence splitting and merging are impossible to model with this naive alignment strategy, but were frequently found to be the issue of sub-par alignments in a manual review of preliminary results. In particular, we also note that there were both cases of several relevant Wikipedia sentences for a single Klexikon sentence (highlighting the importance of a notion of "relevance"), as well as instances of long sentences from Wikipedia splitting into several (non-consecutive) sentences in the Klexikon text.

%For sentence alignments, we identify the most appropriate excerpt in the Wikipedia source for each sentence in the Klexikon article. To measure similarity, we utilize the multilingual version of MiniLM.\footnote{\textcolor{red}{include reference}} To account for sentence compression and omission during the simplification process, we generate sentence-grams of up to three sentences, and decide the final matching based on highest scores.

\subsection{Comparison to Existing Resources}
The only other two German datasets with document alignments are the recent resource by \cite{battisti-etal-2020-corpus}, as well as a smaller version of Klexikon data by \cite{hewett-stede-2021-automatically}. \cite{battisti-etal-2020-corpus} compiled documents from accessibility options on websites. Compared to our dataset, they potentially cover a more heterogeneous set of topics, but only provide alignments for a subset of articles.
As mentioned before, \cite{hewett-stede-2021-automatically} provide additional alignments to MiniKlexikon, and otherwise limit the maximum length of articles, which reduces the number of available alignments between all three resources to 295 documents. Even when considering only the equivalent Klexikon-Wikipedia alignments, there are less than $1{,}000$ documents, with additional constraints to the completeness of the Wikipedia texts.

Concerns raised about the quality of Wikipedia as a resource \cite{xu-etal-2015-problems} mention the problems with sentence alignment, inadequate simplifications, and poor generalization. Our version of the Klexikon dataset partially alleviates these issues:
\begin{enumerate}
	\item We provide document and (automated) sentence alignments, which allows focusing on both summarization and simplification in a joint manner.
	\item Articles for Klexikon are written following stricter guidelines both in their content structure, and we include stricter pre-processing criteria for the Wikipedia articles, resulting in a high-quality collection of text documents.
	\item We provide sufficient training samples for potential neural approaches, by increasing the Klexikon-based resource to almost $2{,}900$ articles.
\end{enumerate}

%\begin{itemize}
%	\item SimpleWiki: Similar in size (after cleaning, I think), but of course no alignment. Several iterations of this one have been published by several groups, e.g., Gurevych, Hwang and Kauchak.
%	\item Geo/Geolino corpus: First German resource of significant size, also several differnt versions. Suffers from the problem that Geo/Geolino doesn't agree to make them public. Also, if I understand it correctly, they have no real alignment in the original dataset, and a narrower focus on the topics that are discussed.
%\end{itemize}

\subsection{Baseline Performance}
\label{sec:baseline}
To quantify the quality of our automatically generated alignments, we investigate the dataset from both a summarization and simplification perspective.

\subsubsection{Summarization}
To verify the suitability of our corpus for \emph{summarization} purposes, we computed several  baselines and compared them to the Klexikon articles as a presumable gold standard summary:
\begin{enumerate}
	\item \textbf{Lead-$3$:} A baseline frequently used in news article summarization, which consists of the first three sentences. In our case, this corresponds to the first three sentences of the Wikipedia article.
	\item \textbf{Lead-$k$:} A related baseline, taking all sentences of the overview section in the Wikipedia article.
	\item \textbf{Full article:} The full Wikipedia article as a reference for the maximum possible vocabulary overlap (this corresponds to ROUGE-1 recall).
	\item \textbf{ROUGE-2 oracle:} As an approximation of the upper limit for extractive summaries on this dataset, we select the sentence maximizing ROUGE-2 F1 scores for each sentence in the Klexikon article and 
	\item \textbf{Luhn:} A simple unsupervised baseline for extractive summaries can be generated by Luhn's algorithm~\cite{luhn-1958-automatic}. We use a target of 25 extracted sentences for each generated summary, which corresponds roughly to the median number of sentences in the Klexikon articles.
	\item \textbf{LexRank S-T:} As a more sophisticated baseline, this approach supplies LexRank~\cite{erkan-radev-2004-lexrank} with embeddings extracted by \texttt{sentence-transformers}~\cite{reimers-gurevych-2019-sentence}\footnote{The same model as mentioned in footnote 7 is used.}~. The length is similarly limited to at most 25 extracted sentences.
	
\end{enumerate}

We use ROUGE \cite{lin-2004-rouge} to gauge summarization quality, which evaluates $n$-gram overlap between system outputs and gold references. In particular, we report F1 scores for ROUGE-1, ROUGE-2 and ROUGE-L.
Results in \Cref{tab:rouge} indicate that our dataset poses a significantly harder challenge compared to performance of baselines on standard summarization corpora, such as CNN/DailyMail \cite{nallapati2017summarunner}, where simple lead-3 baselines obtain extremely high ROUGE scores due to an overly pronounced lead bias.

On our dataset, lead-$3$ likely struggles with the very different output lengths and comparatively low recall scores; the opposite is true for the full article baseline, which does not summarize at all, and therefore scores poorly in terms of precision.
However, the full article baseline obtains a recall score of $77.3\%$ ROUGE-1, implying there is still a sizable vocabulary overlap between the Klexikon and Wikipedia articles.
With proper summarization methods, it is therefore possible to produce decent ROUGE scores, and another indicator of the corpus' suitability to summarization.
Best-suited as a baseline is lead-$k$, which is a decent approximation of the actual target article length. Even so, lead-$k$ is shorter than the corresponding Klexikon articles.
Based on these results, coupled with varying compression levels  between articles (cf. \Cref{fig:stats}~), a high sensitivity to the overall input length seems to be required in order to generate appropriate summaries.

From the extractive summaries generated by unsupervised methods, it becomes obvious that content from sections outside the overview paragraph is beneficial in terms of ROUGE scores, which is a promising distinction from other summarization datasets, especially in German.
Finally, the ROUGE-2 oracle gives insights into the limitations of extractive summarization methods on this dataset. In particular, the differing expressiveness and vocabulary impacts the achievable ROUGE-2 and ROUGE-L scores. It should be noted, however, that the determination of output lengths seems to play a crucial role in the overall balance between precision and recall scores. Given that both unsupervised baselines work with informed choices of the expected summary length, their results should also be taken within the correct context.

\begin{table}
	\centering
	\begin{tabular}{lr|r|r}
		\hline
		\textbf{} & \textbf{R-1} & \textbf{R-2} & \textbf{R-L}\\
		\hline
		Lead-$3$ & $16.95$ & $3.77$ & $9.81$ \\
		Lead-$k$ & $24.87$ & $5.10$ & $12.01$ \\
		Full article & $16.81$ & $4.23$ & $6.95$ \\
		\hline
		ROUGE-2 oracle & $41.85$ & $10.68$ & $16.00$ \\
		\hline
		Luhn & $31.86$ & $5.55$ & $11.57$ \\
		LexRank S-T & $33.90$ & $6.11$ & $12.86$ \\
		
	\end{tabular}
	\caption{Average ROUGE F1 for simple extractive baselines. $95\%$ confidence intervals for all scores differ by less than one point.}
	\label{tab:rouge}
\end{table}

\subsubsection{Simplification}

We further provide different metrics to estimate the level of simplification present in the available documents.
For this, we compute Flesch reading-ease scores \cite{flesch1948new}, specifically an adjusted variation for German \cite{amstad1978verstandlich}.
In addition, we hypothesize that the average sentence length (in tokens), as well as the average number of characters per words are suitable proxies for simplification.
The latter is especially important for German, which is famous for its long compound words. In particular, we limit the word length calculation to "content word classes", i.e., nouns, verbs, adjectives, and adverbs only. \\
To cover lexicographic peculiarities in the data, we estimate the underlying vocabulary.  Notably, the overall texts are quite different in lengths, so an absolute count of distinct tokens would heavily bias the results on Wikipedia.
Instead, we approximate this problem by looking at corpus-specific lemma coverage. By computing a corpus-specific list of the 1000 most frequently occurring lemmas, we are then able to compute what fraction of all used lemmas is contained in this top-1000 list.
A higher percentage likely points to fewer rare words used, and greater reliance on commonly understood words or an overall smaller vocabulary.

Indeed, we find a consistent pattern in our data (cf.~\Cref{tab:simpler}), where Klexikon data indicates simpler language on all our metrics,  which confirms the suitability of our dataset for \emph{simplification} tasks.
We would like to point out the general consensus of the field that heuristics are only scratching the surface of representative readability judgments \cite{chall1958readability}, but still offer a chance for initial exploratory analysis of data suitability.

\begin{table}[t]
	\centering
	\begin{tabular}{lr|r}
		\hline
		\textbf{} & \textbf{Wikipedia} & \textbf{Klexikon} \\
		\hline 
		Avg.~Flesch score & $40.1 \pm 7.3$ & $66.7 \pm 6.0$  \\
		Avg.~sentence length & $22.7 \pm 2.6$ & $13.5 \pm 1.5$ \\
		Avg.~word length & $8.7 \pm 4.0$ & $6.9 \pm 3.0$ \\
		Share of top 1000 lemmas & $68.8$\% & $82.3$\% \\
	\end{tabular}
	\caption{Indicators of simplified target texts: averages for Flesch complexity scores (between 0 to 100; higher scores indicate simpler texts); average sentence length in tokens; average word length in characters (nouns, verbs, adjectives, adverbs); percentage share of occurrences of the top-1000 corpus-specific lemmas.}
	\label{tab:simpler}
\end{table}

%%%%%%%%%%%%%%%%%%%%%%%%%
\section{Conclusion and Future Work}
In this work, we laid out basic requirements for a unified Text Simplification and Summarization framework.
Specifically, we also provided a document-aligned resource of German texts to facilitate future research in this area, and provide quantitative evidence of the suitability of our dataset.
We see the following points as the most critical issues for successful joint models:
\begin{inparaenum}[i)]
	\item Learned sentence relevance and simplification in a joint setting. This can be potentially achieved by modeling sentence alignments similar to existing methods~ \cite{stajner-etal-2018-cats,jiang-etal-2020-neural}, but already during the training of an end-to-end system, instead of a separate pre-processing step.
	\item Implementation of automated evaluation metrics that align both with human judgments of appropriateness for the summary, as well as simplification steps taken. ROUGE, based on $n$-grams, potentially suffers similar shortcomings to BLEU as an evaluation metric, since it fails to capture lexicographic simplifications. Existing simplification metrics, however, are unable to quantize the quality based on much longer source documents.
	\item Extension of current abstractive summarization systems towards lexicographic simplification, potentially in the form of regularization during training.
\end{inparaenum}

%\section{Conclusion}
%
%Thus, a language resource should be cited as \citelanguageresource{Speecon} and \citelanguageresource{EMILLE} .

% \nocite{*}
\section{Bibliographical References}\label{reference}
%\label{main:ref}

\bibliographystyle{lrec2022-bib}
\bibliography{custom}

%\section{Language Resource References}
%\label{lr:ref}
%\bibliographystylelanguageresource{lrec2022-bib}
%\bibliographylanguageresource{languageresource}

\appendix
%%%%%%%%%%%%%%%%%%%%%%%%%%%%%%%%%%%
\section{Experimental Resources and Parameters}
For the evaluation of ROUGE scores, we used the Python implementation provided by Google Research.\footnote{\url{https://github.com/google-research/google-research/tree/master/rouge}}
We replace the original stemmer with Cistem \cite{weissweiler2017developing} to account for appropriate treatment of German tokens.

Flesch complexity scores were computed with the \texttt{textstat} library\footnote{\url{https://github.com/shivam5992/textstat}}, using the function for German. Sentence length in tokens was derived from the tokenization mentioned in the main article.

\section{Data Split}
We additionally present a stratified data split for the corpus, with an approximate 80/10/10 split for training, validation and testing.
For stratification, we represent each pair of source/simplification documents by their respective lengths in number of sentences.
We then divide the coordinate system into a rectangular grid (steps of 100 for Wikipedia article length, step size 10 for Klexikon), and proceed to sample from each grid block according to our pre-defined split (10\% of grid samples are selected for validation, 10\% for testing, and the remaining 80\% for training). When fewer than ten samples are within a block, all samples are added to the training set. This results in a final split of 2350 training pairs, and 274 samples each for validation and testing. The split is available through the previously mentioned Github repository.

\end{document}